\pdfoutput=1

\documentclass[11pt]{article}

\usepackage{ACL2023}

\usepackage{times}
\usepackage{latexsym}

\usepackage[T1]{fontenc}

\usepackage[utf8]{inputenc}

\usepackage{microtype}

\usepackage{inconsolata}

\usepackage{amsmath}      
\usepackage{amssymb}      
\usepackage{mathdots}     
\usepackage{array}

\usepackage{graphicx}     
\usepackage{booktabs}     
\usepackage{multirow}     
\usepackage{adjustbox}    
\usepackage{subcaption}   

\usepackage{algpseudocode}  
\usepackage[ruled,vlined,linesnumbered]{algorithm2e} 

\usepackage{pifont}      
\usepackage{xspace}      

\usepackage{xcolor}      
\usepackage[skins]{tcolorbox} 

\usepackage{enumitem}

\title{Mapping the Minds of LLMs: A Graph-Based Analysis of Reasoning LLM}

\author{Zhen Xiong$^\lambda$ \quad Yujun Cai$^\ddagger$ \quad Zhecheng Li$^\delta$ \quad Yiwei Wang$^\dagger$ \\
$^\lambda$ University of Southern California \quad $^\ddagger$ The University of Queensland \\ $^\delta$ University of California, San Diego \quad $^\dagger$ University of California, Merced \\
  \texttt{xiongzhe@usc.edu} \\}

\begin{document}

\maketitle

\begin{abstract}
Recent advances in test-time scaling have enabled Large Language Models (LLMs) to display sophisticated reasoning abilities via extended Chain-of-Thought (CoT) generation. Despite their potential, these Reasoning LLMs (RLMs) often demonstrate counterintuitive and unstable behaviors, such as performance degradation under few-shot prompting, that challenge our current understanding of RLMs. In this work, we introduce a unified graph-based analytical framework for better modeling the reasoning processes of RLMs. Our method first clusters long, verbose CoT outputs into semantically coherent reasoning steps, then constructs directed reasoning graphs to capture contextual and logical dependencies among these steps. Through comprehensive analysis across models and prompting regimes, we reveal that structural properties, such as exploration density, branching, and convergence ratios, strongly correlate with reasoning accuracy. Our findings demonstrate how prompting strategies substantially reshape the internal reasoning structure of RLMs, directly affecting task outcomes. The proposed framework not only enables quantitative evaluation of reasoning quality beyond conventional metrics but also provides practical insights for prompt engineering and the cognitive analysis of LLMs. Code and resources will be released to facilitate future research in this direction.
\end{abstract}

\section{Introduction}

Recent LLMs equipped with test-time scaling capabilities, such as OpenAI's o-series~\cite{openai2024o1, openai2025o3o4}, DeepSeek-R1~\cite{deepseek2025r1}, and Gemini-2.5~\cite{google2025gemini}, employ a system II, \textit{think-slow-before-answer}, pipeline that transforms how these models approach complex problems during test time. Rather than producing outputs directly after the input with normally limited token length, these reasoning models engage in explicit and free extended reasoning through Chain-of-Thought ~\cite{cot2022} mechanisms. This innovation enables reasoning models to decompose intricate challenges in various domains, explore multiple possible solutions, and self-assess intermediate conclusions before synthesizing final responses during their extended inference time. In general, these reasoning models currently outperform conventional LLMs on various types of benchmarks, which require advanced math~\cite{patel2024aime} and coding~\cite{jimenez2024swebench} capability.

Despite these promising advancements, reasoning models exhibit undesire~\cite{chen2024donotthink} and unstable~\cite{yang2025thinking} behaviors that challenge the established understanding of large language models. One of the particularly striking phenomena is the performance degradation associated with few-shot learning, which in most cases improves the performance of conventional LLMs. Recent technique reports also documented that these RLMs are somehow more sensitive to prompts~\cite{deepseek2025r1}. We believe these existing unclear behaviors of RLM call for deeper investigations into how RLMs operate and reason. 

\begin{figure}[bt]
    \centering
    \includegraphics[width=1.0\linewidth]{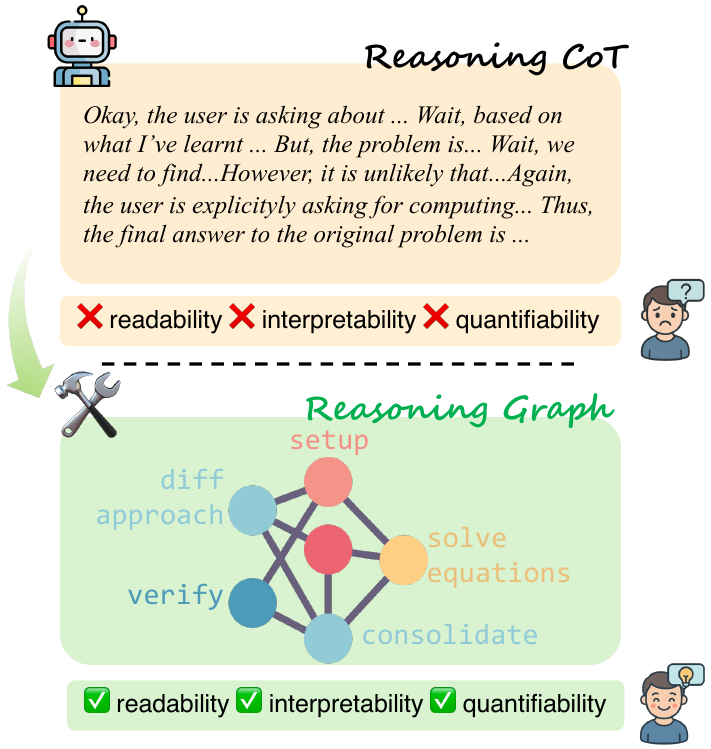}
    \caption{A conceptual overview of our framework for modeling the long reasoning CoT with a graph structure. This graph-based representation enables stronger readability for human researchers, systematic interpretability of the global structure, and quantifiable graph metrics for in-depth analysis.}
    \label{fig:cot_to_graph}
\end{figure}

Our research proposes a novel framework to trace the reasoning processes from a graph perspective. While some work has previously examined the correlation between the quantity of reasoning tokens and RLM's accuracy~\cite{sui2025stop, relationship2025, yang2025thinking}, our approach goes beyond the token perspective and focuses on the semantic organization of the model's reasoning processes. Specifically, our analytical frameworks first cluster raw and verbose reasoning tokens into coherent logic steps and then map their inter-dependencies as a graph, enabling globally semantical insights into how reasoning models reason at a higher level (Figure~\ref{fig:cot_to_graph}). After a comprehensive analysis of derived reasoning graphs, we identify specific quantifiable features that are associated with advanced reasoning behavior, which is often linked to higher problem-solving performance.

To summarize, our contributions in this paper include:
\vspace{-0.3em}
\begin{itemize}
    \item a novel reasoning-graph toolkit that converts natural language long Chain-of-Thought into analyzable graph structures, enabling quantification of reasoning through topological and semantic metrics.
    \item comprehensive analysis of how different prompting strategies may influence reasoning LLMs, establishing quantitative boundaries for prompt engineering optimization.
    \item quantifiable indicators of reasoning quality beyond task accuracy, providing a higher-level cognitive understanding of reasoning in LLMs.
\end{itemize}

\section{Related Works}

\paragraph{Test-Time Scaling} Similar to human dual-processing hypothesis of the mind~\cite{da2023system}, augmenting the computational budget at test-time has been shown to substantially enhance the reasoning capabilities of large language models (LLMs)~\cite{openai2024o1, openai2025o3o4, google2025gemini, anthropic2025claude}. These reasoning LLMs (RLMs) show highly advanced self-reflection, backtracking, and cross-validation behavior during the extended chain-of-thought (CoT) responses, enabling them to tackle intricate reasoning challenges and outperform previous conventional base LLMs~\cite{li2025system, chen2025towards}.

\paragraph{Few-Shot Learning} Few-shot prompting once emerged as a crucial technique for enhancing the performance and adaptability of large language models (LLMs) by providing limited yet highly informative demonstrations~\cite{song2023fewshot}. In detail, it leverages a minimal number of illustrative examples embedded directly into the input context, enabling models to rapidly generalize across diverse tasks without explicit parameter updates~\cite{language2020}. However, many researchers and practitioners have reported that few-shot prompting could instead degrade the model's performance~\cite{deepseek2025r1}, signaling the instability of current reasoning LLMs. In this paper, we will examine the impact of zero/few-shot prompting on RLM's reasoning, assessing both the quality of internal reasoning and overall performance in in-context learning scenarios. Provide more valuable insights for future prompt engineering and model optimization.

\paragraph{Long CoT Analysis} Some previous studies have identified a negative relationship between an RLM's accuracy and the number of reasoning tokens it generates~\cite{relationship2025, yang2025thinking}. However, their analyses of RLMs mainly relied on a one-dimensional metric: the length of CoT token sequences. It still remains unclear and counterintuitive why even longer system II thinking could lead to performance degradation, suggesting a gap in our understanding of how RLMs work in general. In this work, we introduce a comprehensive structured framework to formulate the chain-of-thought process, offering deeper insights into the underlying reasoning behavior. 

\section{Constructing Reasoning Graph from Raw Reasoning Tokens}
\label{sec:graph-construction}

\begin{figure*}[tb]
    \centering
    \includegraphics[width=1.0\textwidth]{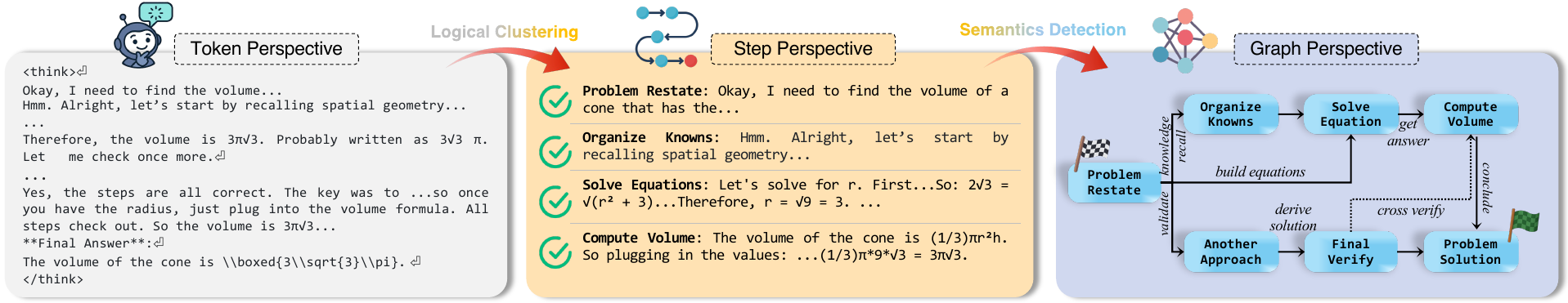}
    \caption{Our pipeline for building the graph structure from reasoning large language models' output. Starting from raw \textbf{token perspective}, we first use "\textbackslash n\textbackslash n" as natural delimiters to split the raw reasoning tokens into an ordered list of reasoning units. Then we perform logical clustering to combine logically cohesive reasoning units into a reasoning step (\textit{node}), shifting into intermediate \textbf{step perspective}. Lastly, we detect semantical relationships (\textit{edge}) between steps (\textit{node}) to reveal the high-level \textbf{graph perspective} from reasoning LLM's output.}
    \label{fig:method}
\end{figure*}

Given more computational resources at inference time, Reasoning Language Models (RLMs) can autonomously explore feasible solutions, perform cross-validations, actively access intermediate steps, and synthesize consolidations through extended chain-of-thought tokens. This critical feature allows RLMs to fully release their internal reasoning potential under sophisticated challenges. However, current test-time scaling is also a double-edged sword: as models are encouraged to elaborate and reflect, their behavior often becomes unreliable and less predictable. 

Counterintuitively, the “thinking out loud” style of RLM should, in theory, offer richer data for LLM interpretability research and help us understand how LLM actually reasons. Yet, to the best of our knowledge, there is still a lack of effective methods to systematically analyze and model the semantic content of RLM-generated reasoning tokens. To fill this gap, we propose a novel, structured approach for representing and dissecting the reasoning process of RLMs.

It is widely acknowledged that RLMs tend to generate complex, branching chains of thought. This pattern closely mirrors the way humans reason: rather than following a strictly linear path, our thinking often leaps between ideas, drawing on contextual cues, connecting prior knowledge and memories, searching for potential solutions, and constantly checking for errors along the way. It is precisely this interplay of multiple analytical paths that allows us to synthesize a coherent conclusion. Inspired by this convoluted human \textit{mind map}, we propose a unified, graph-based framework (Figure~\ref{fig:method}) to model the structure of RLM outputs.

\subsection{Graph Formalization}
We can formally define the reasoning graph $G = (V, A)$ with the following components:

\begin{itemize}
    \item $V = \{s_1, s_2, ..., s_n\}$: An ordered list of vertices representing semantically clustered reasoning steps.
    \item $A \in \{-1, 0, 1\}^{n \times n}$: An adjacency matrix representing the ternary logical relationship between reasoning steps.
\end{itemize}

In the remaining part of this section, we will first introduce a method for clustering long and verbose reasoning traces into discrete, semantically coherent reasoning steps, each of which will serve as a \textbf{node} in a reasoning graph (Section~\ref{sec:build-node}). We then describe how to extract semantic dependencies between these steps to form the \textbf{edges} of the reasoning graph ((Section~\ref{sec:build-edge})). This reasoning graph construction method will be used in subsequent sections as a key tool for quantitatively analyzing RLM's behavior.

\subsection{Clustering Raw Tokens into Discrete Reasoning Steps}
\label{sec:build-node}

Long chain-of-thought (CoT) sequences generated by RLMs often span thousands of tokens. While these detailed traces offer rich insight into the model's reasoning process, their length and fragmented nature present challenges for systematic analysis. A common pre-processing strategy is to segment the output based on explicit delimiters: RLMs frequently insert the token ``\texttt{\textbackslash n\textbackslash n}'' to denote boundaries between successive thoughts. Let $T = (t_1, t_2, \ldots, t_N)$ represent the generated token sequence and use $\mathcal{D} = \texttt{"\textbackslash n\textbackslash n"}$ as the delimiter. We thus obtain an initial partition into \emph{reasoning units}:
\[
U = (u_1, u_2, \ldots, u_M),
\]
where each $u_i$ is a contiguous subsequence bounded by delimiters, i.e., $u_i = (t_{s_i}, \ldots, t_{e_i})$ and $t_{e_i+1} = \mathcal{D}$.

Despite its simplicity, delimiter-based segmentation has two fundamental limitations. For complex tasks such as advanced mathematical reasoning or code generation, $M$ can be excessively large, resulting in an unwieldy number of fragmented units that hinder semantic analysis and dependency extraction. Moreover, the model's stylistic tendency to insert delimiters frequently can lead to reasoning units that are too fine-grained, often lacking coherent context for standalone analysis.

\paragraph{Context-aware Logical Units Clustering.}

\begin{figure}[tb]
    \centering
\begin{tcolorbox}[width=\linewidth, fonttitle = \small\bfseries, title=$\mathcal{P}_{\text{clu}}$ \quad Context-aware Logical Units Clustering ,colframe=gray!2!black,colback=gray!2!white,boxrule=1pt,boxsep=0pt,left=5pt,right=5pt,fontupper=\footnotesize, halign title = flush center]
\textbf{Instruction:} 

You are given a sequence of \textcolor{red}{reasoning units}...

[\textit{Logical Units Template}]

\quad\\
Your task is to \textcolor{red}{cluster consecutive} reasoning units that are \textcolor{red}{semantically connected}...\\

Expected Output Format: [\textit{Output Guideline}]
\tcbline
\textbf{\color[RGB]{0,0,0}{Output}:} \{"$s_0$": \{"title":..., "content":...\},...\}

\end{tcolorbox}

\caption{Our abbreviated prompt template to guide LLM to cluster reasoning units into logical cohesive reasoning steps. For detailed \textit{Logical Units Template} \& \textit{Output Guideline} see Appendix~\ref{appendix:prompts}.}
\label{prompt:clustering}
\end{figure}

To address these challenges, we introduce a context-aware logical units clustering procedure that aggregates semantically related reasoning units into higher-level \emph{reasoning steps}. Specifically, we leverage a large language model (LLM) to sample possible clusterings under decoding temperatures $\tau_r$ conditioned on a carefully designed prompt template $\mathcal{P}_{\text{clu}}$ (see Figure~\ref{prompt:clustering}) $\mathcal{P}_{\text{clu}}$:
\[
S = (s_1, s_2, \ldots, s_K) \sim P_{\text{LLM}}(S \mid \mathcal{P}_{\text{clu}},\, U; \tau_{r}),
\]
where each $s_j$ is ideally formed by concatenating adjacent $u_i$ meeting a semantic affinity criterion, with $K \ll M$. This aggregation aims to ensure that each reasoning step $s_j$ provides sufficient context for downstream analysis while maintaining a manageable total number of segments.

Yet, given the generative nature of LLMs, repeated invocations of the clustering prompt do not guarantee an identical clustering. Rather than treating this variability as noise, we further harness it through a further ensemble sampling and selection approach to identify the most coherent clustering.

\paragraph{Ensemble Sampling}
To capture the full range of possible clustering, we generate an ensemble of $B$ candidate instances of clustering by independently sampling $P_{\text{LLM}}(S \mid \mathcal{P}_{\text{clu}},\, U)$ from  with varied random seeds and decoding temperatures:
\begin{align*}
\mathcal{C} &= \{C^{(1)}, C^{(2)}, \ldots, C^{(B)}\} \\
    C^{(b)} &= (s^{(b)}_{1}, \ldots, s^{(b)}_{K_b})
\end{align*}
Each $C^{(b)}$ is a candidate clustering of the original reasoning units.

To objectively compare these candidates, we introduce a weighted average quality score that evaluates key properties of each clustering. In general, we compute
\[
F(C^{(b)}) = \sum_{\ell=1}^{L} w_\ell\,\phi_\ell(C^{(b)}),
\]
where each $\phi_\ell$ is a quantitative metric and the weights $w_\ell$ are non-negative and sum to one. In our experiments, we consider three intuitive criteria:

\begin{itemize}[left=0pt,align=left,labelwidth=3em,labelsep=0.5em]
    \item[\textit{Criteria 1}] \textbf{Intra-step coherence.}
    \[
    \phi_{\text{ic}}(C^{(b)}) =
    \frac{1}{K_b}
    \sum_{j}
    {
      \frac{2\sum_{\substack{u<v \in s^{(b)}_{j}}}
      \mathrm{cos}(\mathbf{e}_u, \mathbf{e}_v)}{|s^{(b)}_{j}|(|s^{(b)}_{j}|-1)}
    }
    \]
    which rewards semantic similarity among units within each reasoning step.

    \item[\textit{Criteria 2}] \textbf{Step-to-step separation.}
    \[
    \phi_{\text{sep}}(C^{(b)}) =
    \frac{1}{K_b-1} \sum_{j=1}^{K_b-1}
    \left[1 - \mathrm{cos}(\mathbf{e}_{s^{(b)}_{j}}, \mathbf{e}_{s^{(b)}_{j+1}})\right]
    \]
    which encourages semantic distinctiveness between adjacent reasoning steps.

    \item[\textit{Criteria 3}] \textbf{Length regularity.}
    \[
    \phi_{\text{len}}(C^{(b)}) =
    1 - \left|\frac{\frac{1}{K_b}\sum_{j} |s^{(b)}_{j}|}{\mu_{\text{ref}}} - 1\right|
    \]
    where $\mu_{\text{ref}}$ is a referenced average step length derived from the original CoT structure (see Appendix~\ref{appendix:mu_ref} for details), penalizing pathological segmentations that are too short or too long.
\end{itemize}

Here, $\mathbf{e}_\cdot$ denotes a sentence embedding vector derived from a pretrained encoder. 

The final segmentation is selected as the candidate with the highest composite score:
\[
V := C^{*} = \arg\max_{C \in \mathcal{C}} F(C).
\]
This ensemble–scoring framework offers a flexible yet principled way to select coherent and analyzable reasoning step from all plausible instances of clustering generated by the LLM. The complete clustering and selection algorithm is summarized in Appendix~\ref{appendix:algorithm}. 

For all subsequent analyses, each $s_j$ in the selected $C^{*}$ is treated as a node in our reasoning graph, providing a compact yet semantically rich foundation for structural and dependency analysis.

\subsection{Extracting Inter-Dependencies between Reasoning Steps}
\label{sec:build-edge}

our next objective is to construct a directed semantic graph $G=(V, E)$, where each edge $(i,j) \in E$ represents an inferred relationship—such as support or contradiction—between step $s_i$ and step $s_j$. To achieve this, we propose a rejection sampling-based semantic detection procedure that fuses global predictions from multiple LLM samplings.

\paragraph{Adjacency Matrix Sampling.}

\begin{figure}
\centering
\begin{tcolorbox}[width=\linewidth, fonttitle = \small\bfseries, title=$\mathcal{P}_{\text{sem}}$ \quad Adjacency Matrix Sampling,colframe=gray!2!black,colback=gray!2!white,boxrule=1pt,boxsep=0pt,left=5pt,right=5pt,fontupper=\footnotesize, halign title = flush center]
\textbf{Instruction:}

Given an ordered sequence of \textcolor{red}{$K$ reasoning steps},...

[\textit{Logical Steps Template}]

\quad\\
...Your task is to decide whether step\,$i$ \textcolor{red}{supports},
\textcolor{red}{contradicts}, or is \textcolor{red}{independent} of step\,$j$...\\

Expected Output Format: [\textit{Output Guideline}]

\tcbline
\textbf{\color[RGB]{0,0,0}{Output}}: \{($s_0$, $s_1$):..., ($s_0$, $s_2$): ..., ($s_1$, $s_0$): ...\}
\end{tcolorbox}

\caption{Our abbreviated prompt template to detect semantical relationship between two different reasoning steps. For detailed input/output template and intuition behind the instruction, see Appendix~\ref{appendix:prompts}}
\label{prompt:detection}
\end{figure}

We first obtain diverse global views of step-wise dependencies by repeatedly prompting the LLM with a structured template $\mathcal{P}_{\text{sem}}$ (see Figure~\ref{prompt:clustering}). Each prompt presents the entire ordered set of reasoning steps and requests predictions for every ordered pair $(i, j)$, $i < j$. The model outputs a full adjacency matrix:
\begin{align*}
A^{(r)} &\sim P_{\text{LLM}}\left(A \mid \mathcal{P}_{\text{sem}}, V; \tau_r\right) \\
A^{(r)} &\in \{-1, 0, 1\}^{K \times K},
\end{align*}
where $A_{ij}^{(r)} = 1$ (support), $-1$ (contradict), or $0$ (independent). We repeat this sampling $R$ times, varying the decoding temperature $\tau_r$ to enhance diversity. This strategy ensures the LLM can leverage the full context for globally consistent predictions while capturing uncertainty.

\paragraph{Adaptive Edge-wise Probability Estimation.}
For each possible edge $(i, j)$, we aggregate predictions across the $R$ samples to estimate the empirical probability of each relation:
\[
\hat{p}_{ij}(c) = \frac{1}{R} \sum_{r=1}^R \mathbf{1}\left[A_{ij}^{(r)} = c\right], \qquad c \in \{-1, 0, 1\}.
\]
These aggregated probabilities provide a measure of confidence for the existence and type of each possible semantic relation.

Rather than relying on a fixed number $R$ of sampled adjacencies, we fuse information across all samples for a robust final graph construction. Sampling continues until the estimated probabilities for all edges reach a specified confidence level. Specifically, for each edge, we compute the pooled standard error:
\[
\mathrm{SE}_{ij} = 
\sqrt{
    \frac{1}{R}\sum_{l\in\{-1, +1\}}\hat p_{ij}(l)\bigl(1-\hat p_{ij}(l)\bigr)
}
\]
Here, we explicitly omitted ($l=0$) case to simplify the estimator while preserving the accuracy required for the adaptive stopping criterion (see Appendix~\ref{appendix:stderr}). The process halts once $\max_{i<j} \mathrm{SE}_{ij} \leq \varepsilon$ \footnote{in practice $\varepsilon\!=\!0.05$ suffices}, or a hard cap $R_{\max}$ is reached. This guarantees that our edge probability estimations are statistically reliable before moving to the next phase.

With reliable probability estimates, we next construct the final adjacency matrix via a consensus rule. For each pair $(i, j)$, we define the signed confidence:
\[
w_{ij} := \hat p_{ij}(+1) - \hat p_{ij}(-1), \quad w_{ij} \in [-1, 1].
\]
We then apply a dual-threshold criterion:
\[
A_{ij} =
\begin{cases}
+1,& \text{if } w_{ij} \ge \tau_{\text{pos}} \\[0.2em]
-1,& \text{if } w_{ij} \le -\tau_{\text{neg}} \\[0.2em]
0,& \text{otherwise,}
\end{cases}
\qquad
A_{ji} = -A_{ij},
\]
where $\tau_{\text{pos}}$ and $\tau_{\text{neg}}$ can be tuned
independently\footnote{We typically set
$\tau_{\text{pos}}=0.4,\;\tau_{\text{neg}}=0.3$ to reflect the empirical
imbalance between supporting and contradicting links.}.

The resulting weighted adjacency
$W=\bigl[w_{ij}\bigr]_{i,j}$ is preserved for future analysis, while we pay attention to the hard-thresholded
$A=\operatorname{sign}(W)\odot \mathbf{1}[|W|\ge\tau]$, which serves as the
binary backbone for structural analysis. We also include a complete adjacency matrix sampling and adaptive edge estimation algorithm in Appendix~\ref{appendix:algorithm}.

To sum up, the pipeline described in this section provides a principled and systematic framework for converting raw reasoning traces from RLMs into interpretable graph structures. 
This unified graph representation serves as a powerful analytical tool, enabling fine-grained examination of how RLMs organize, connect, and validate intermediate inferences. In the following sections, we will leverage this reasoning graph formalism to quantitatively analyze the internal reasoning dynamics and decision-making behaviors of advanced RLMs.

\section{Reveal Cognitive Behavior of RLM with Reasoning Graph}

Existing analysis of Reasoning LLMs (RLMs) have primarily relied on performance-based metrics such as accuracy or token-level statistics like reasoning length. While these measures offer a coarse understanding of model behavior, they fail to capture the complex and dynamic structure exposed by RLM's output.

In this section, we propose to move beyond token-level perspectives and instead leverage the reasoning graph constructed in Section~\ref{sec:graph-construction} as an effective medium for cognitive analysis. By representing the model’s chain-of-thought as a graph of semantically coherent reasoning steps (nodes) and their directed relationships (edges), we can systematically quantify the structure, flexibility, and effectiveness of model reasoning. This shift enables us to answer deeper questions about how RLMs organize, explore, and consolidate information during problem problem-solving process. Figure~\ref{fig:method} provides a concrete example of RLM's reasoning process for solving a spatial geometric problem.

All implementation details are included in Appendix~\ref{appendix:impl}.


\subsection{Graph-Based Metrics for Quantifying Model Reasoning}

To systematically analyze the cognitive organization of RLM reasoning, we introduce several graph-based metrics, each designed to capture a distinct structural aspect of the reasoning process.

\paragraph{Exploration Density ($\rho_E$):} Measures the overall connectivity among reasoning steps, reflecting the breadth of the model’s exploration. 

\begin{align*}
\rho_E(G) = \frac{|E|}{|V|(|V|-1)}
\end{align*}

Higher values indicate denser intra-reasoning-step correlations.

\paragraph{Branching Ratio ($\gamma_B$):} Quantifies the diversity of alternative reasoning paths, capturing the model’s capacity for exploring parallel ideas and diverse solutions.

\begin{align*}
\gamma_B(G) = \frac{|\{ s \in V \mid \text{d}^-(s) > 1\}|}{|V|}
\end{align*}
where $d^-(s)$ is the out-degree of node $s$.

\paragraph{Convergence Ratio ($\gamma_C$)}:
Captures the extent to which the model integrates multiple reasoning threads into unified conclusions, indicating its ability to synthesize disparate ideas.
\begin{align}
\gamma_C(G) = \frac{|\{ s \in V \mid \text{d}^+(s) > 1\}|}{|V|}
\end{align}
where $d^+(s)$ is the in-degree of node $s$.

\paragraph{Linearity ($\ell$):}
Represents the prevalence of strictly sequential reasoning, measuring the fraction of nodes with degree greater than one.
\begin{align}
\ell(G) = 1-\frac{|\{ s \in V \mid d(s) > 2 \}|}{|V|}
\end{align}
where $d(s)$ is the total degree of node $s$.

Collectively, these metrics offer a comprehensive high-level view of the reasoning graph structure. They allow us to quantify not only how much the model reasons, but how it organizes its thinking through branching exploration, convergence, or rigid linearity. In the following analyses, we will show how these quantities are directly related to and influence model performance.

\subsection{Impact of Prompting Paradigms on Reasoning Structure}

Having established our graph-based metrics, we next investigate how different prompting styles shape the internal reasoning structure of RLMs. We focus on three few-shot demonstration styles: \texttt{Minimal}, \texttt{Concise}, and \texttt{Explanatory}. (see Appendix~\ref{appendix:few_shot_style} for detailed definitions). Each style provides a distinct richness of context, ranging from bare problem–answer pairs to human-like concise reasoning and extended, self-reflective chains generated by the model itself.

\begin{figure}[tb]
    \centering
    \includegraphics[width=0.9\linewidth]{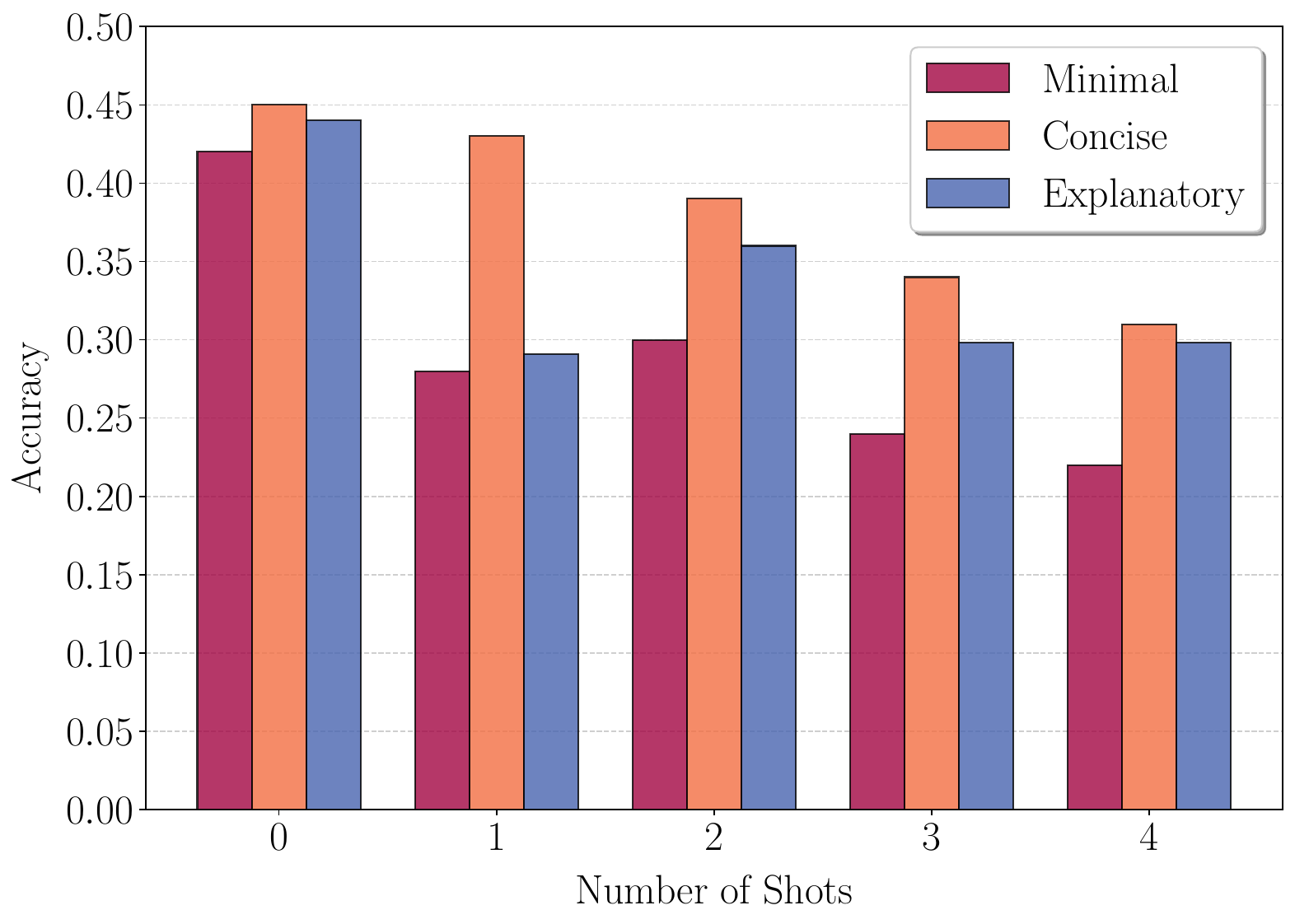}
    \caption{Few-shot prompting accuracy on GPQA-Diamond$^*$ dataset using reasoning Qwen-7B (distilled from DeepSeek-R1). The accuracy drops dramatically with respect to the increasing number of examples within the prompt.}
    \label{fig:acc_drop}
\end{figure}

\paragraph{Prompting Style Modulates RLM Performance.}
Our results reveal a consistent and striking trend: increasing the number of in-context examples leads to a monotonic decline in accuracy, regardless of demonstration style (Figure~\ref{fig:acc_drop}). However, the severity of this decline is strongly related to the structure and verbosity of the provided exemplars, with \texttt{Minial} being the most RLM-unfriendly prompting style.

\begin{figure}[b]
    \centering
    \includegraphics[width=0.9\linewidth]{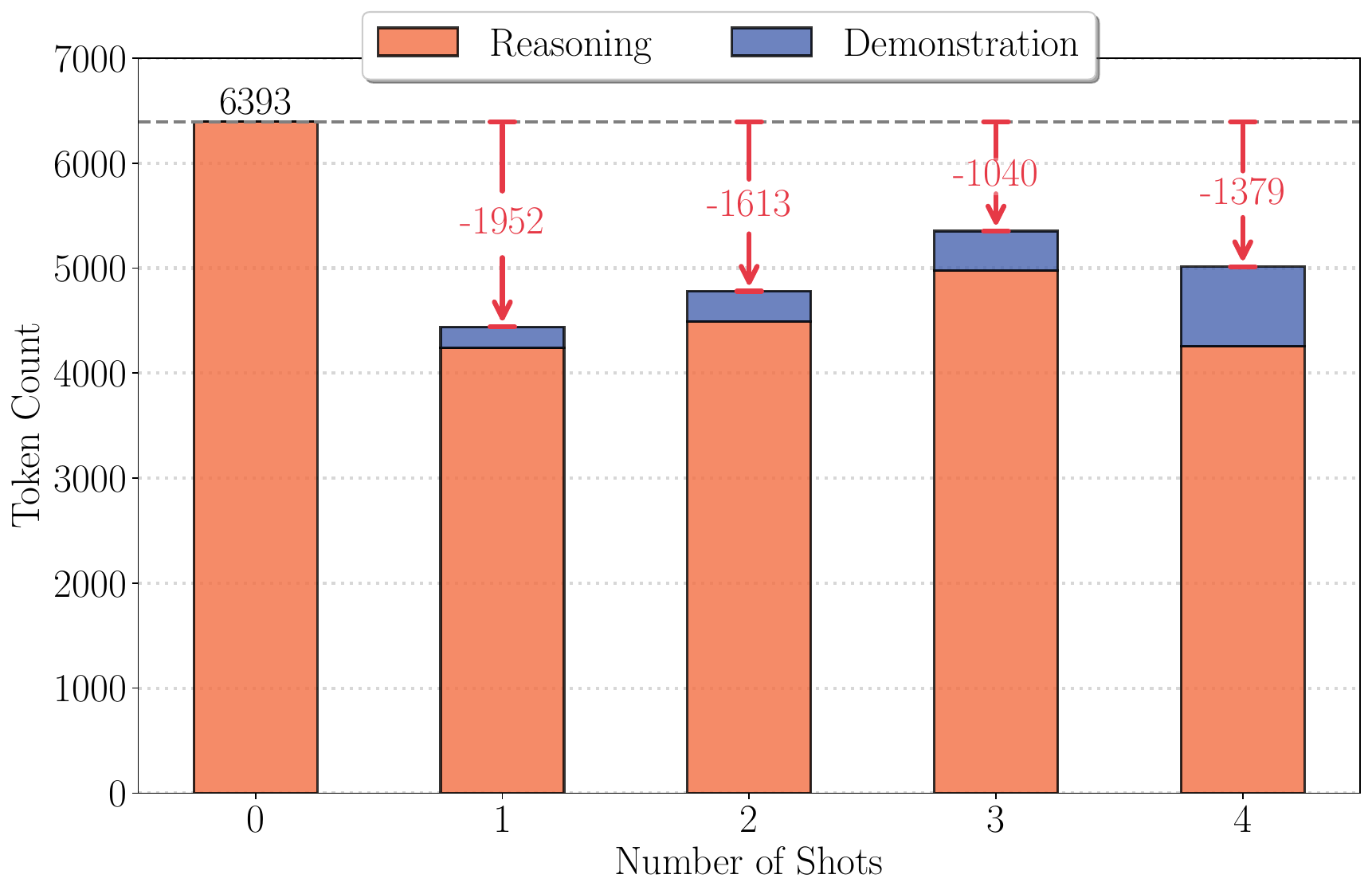}
    \caption{Average number of tokens under different numbers of shots with \texttt{explanatary} few-shot style. Few-shot prompting leads to significantly fewer reasoning tokens compared with zero-shot prompting.}
    \label{fig:total_tokens}
\end{figure}

While there is a hypothesis explaining that few-shot prompting leads to a reduction in total length of I/O tokens  (Figure~\ref{fig:total_tokens}), raw length alone does not fully explain the loss of reasoning effectiveness. Instead, these phenomena call for deeper understanding and explanations for the observed degradation in model performance.

\begin{figure}[tb]
    \centering
    \includegraphics[width=\linewidth]{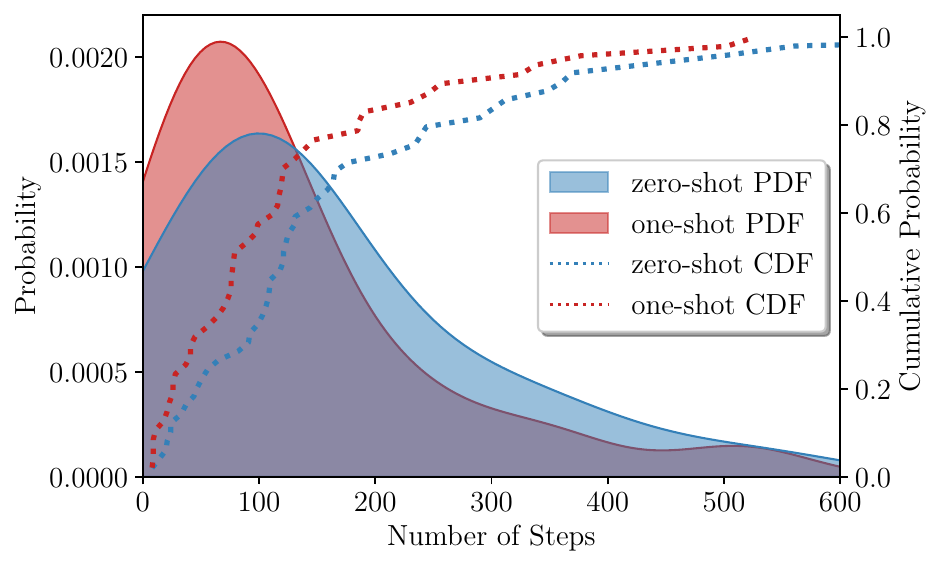}
    \caption{Distribution of reasoning step counts under zero-shot and one-shot prompting. The inclusion of a single demonstration in one-shot settings causes a pronounced shift in the distribution compared to zero-shot, highlighting the sensitivity of RLM to prompt design.}
    \label{fig:step_cdf}
\end{figure}

\begin{figure}[tb]
    \centering
    \begin{subfigure}[t]{0.48\linewidth}
        \includegraphics[width=\linewidth]{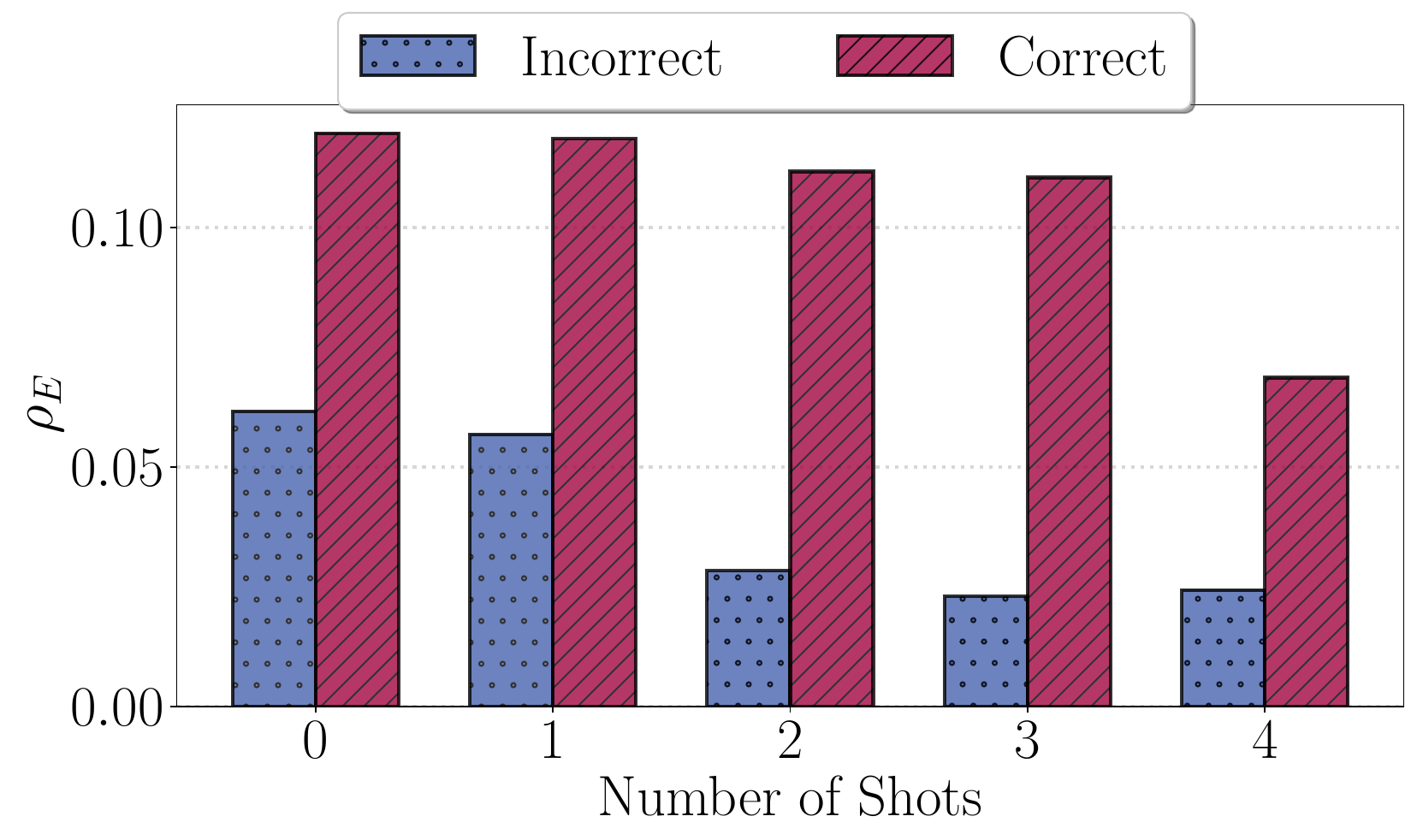}
        \caption{Exploration Density $\rho_E$}
        \label{fig:graph_stats_density_E}
    \end{subfigure}
    \hfill
    \begin{subfigure}[t]{0.48\linewidth}
        \includegraphics[width=\linewidth]{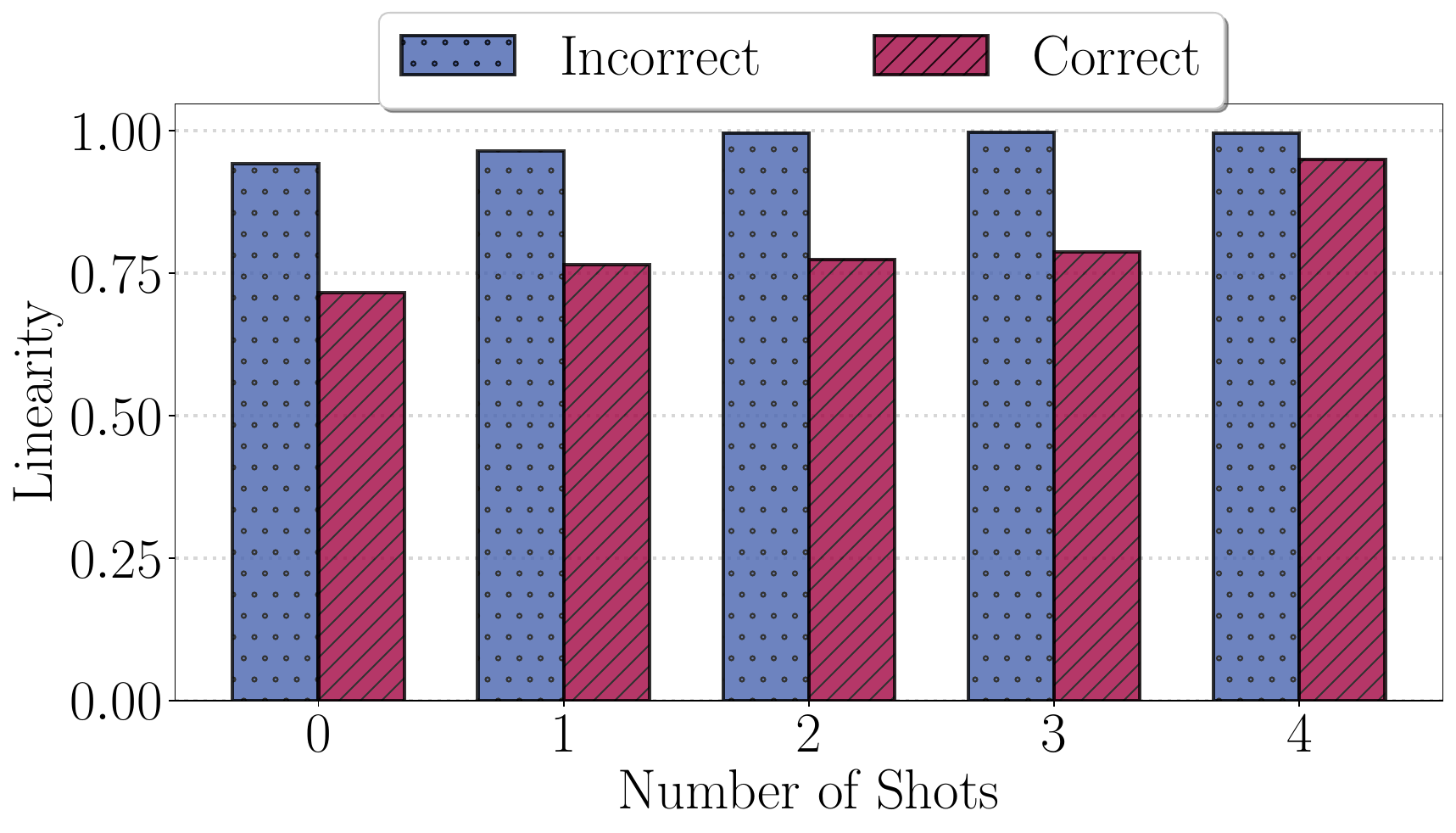}
        \caption{Linearity $\ell$}
        \label{fig:graph_stats_linearity}
    \end{subfigure}
    
    \begin{subfigure}[t]{0.48\linewidth}
        \includegraphics[width=\linewidth]{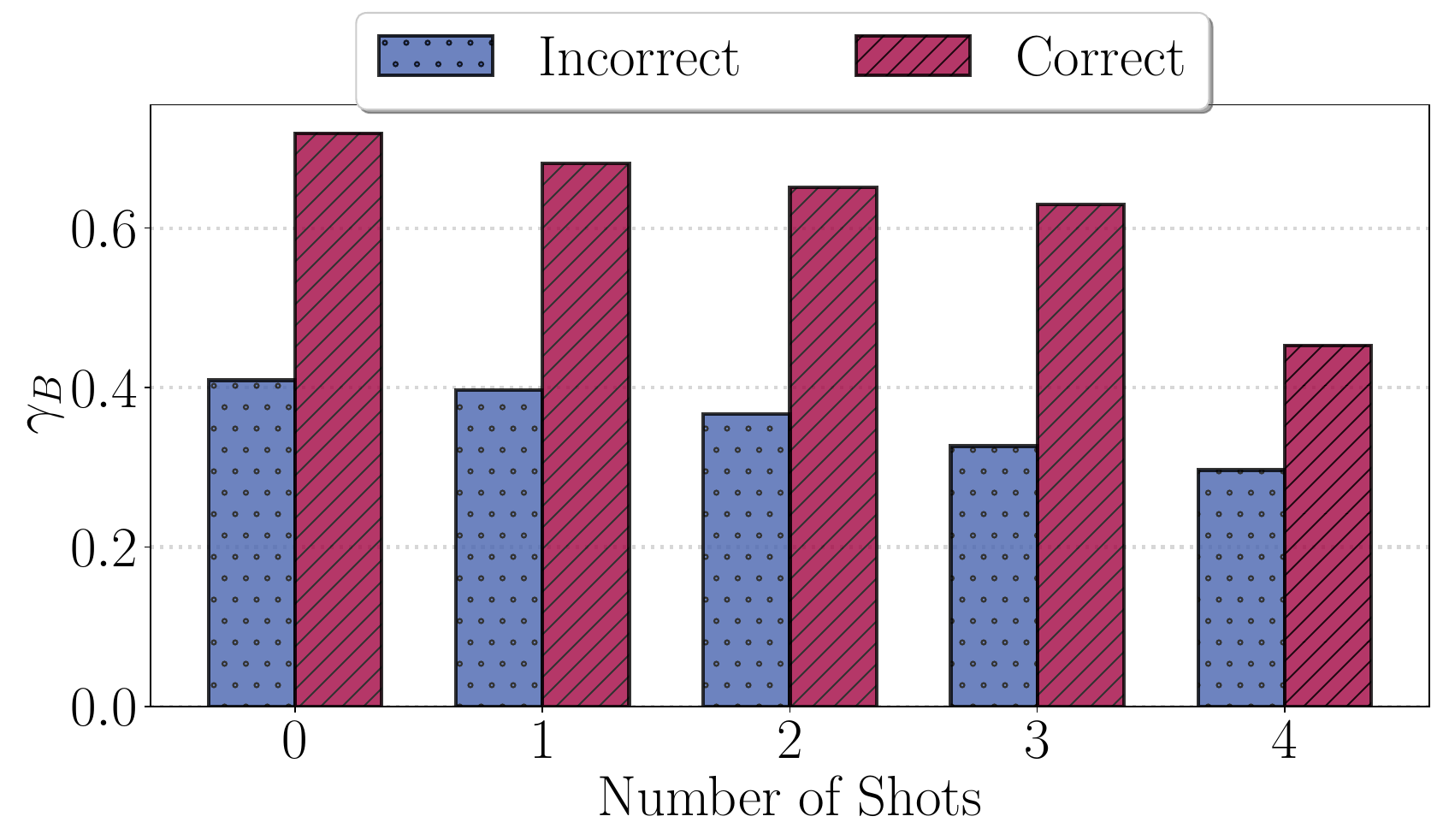}
        \caption{Branching Ratio $\gamma_B$}
        \label{fig:graph_stats_branching}
    \end{subfigure}
    \hfill
    \begin{subfigure}[t]{0.48\linewidth}
        \includegraphics[width=\linewidth]{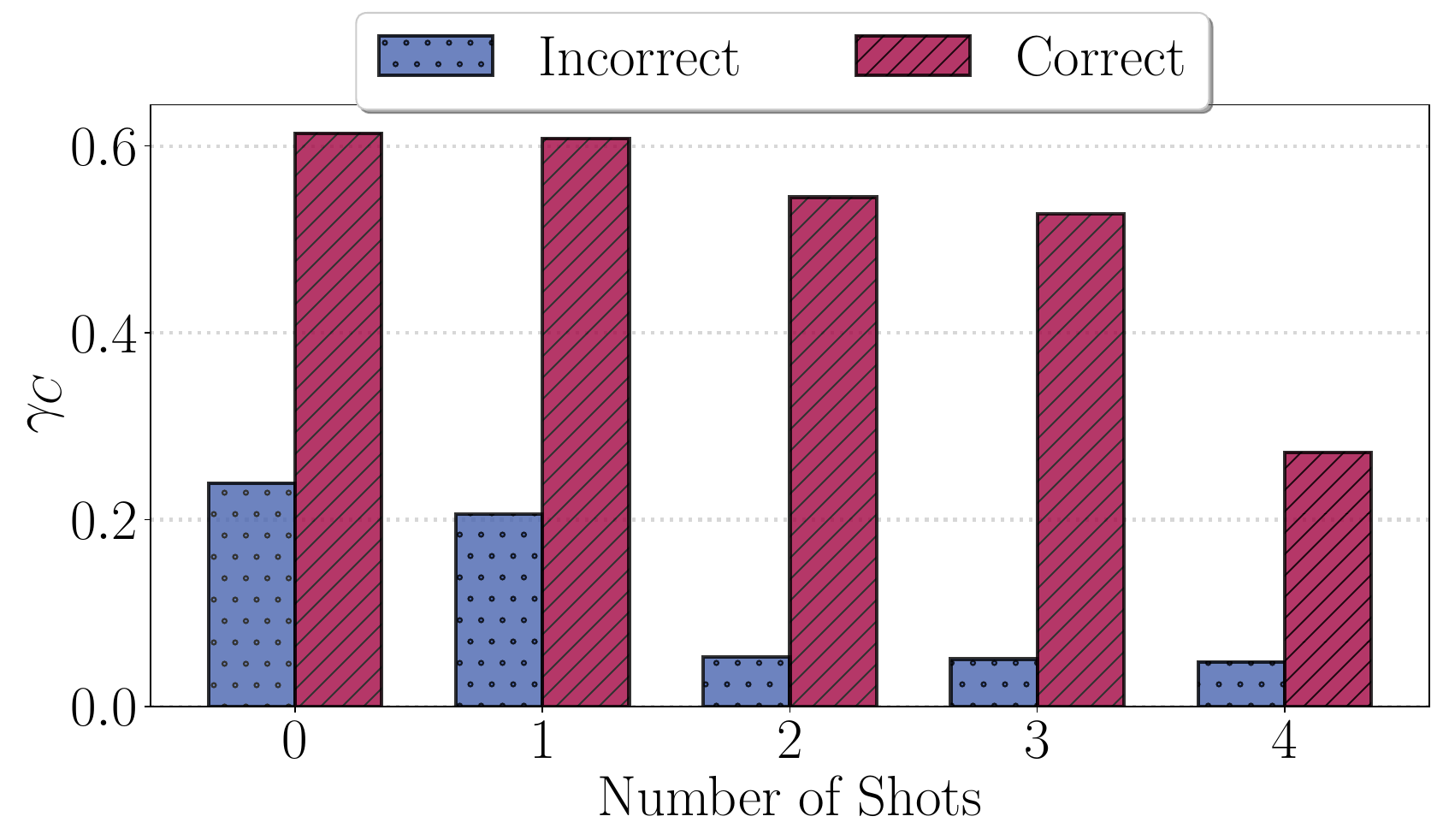}
        \caption{Convergence Ratio $\gamma_C$}
        \label{fig:graph_stats_convergence}
    \end{subfigure}
    \caption{Different metrics of reasoning graph given different numbers of few-shot examples.}
    \label{fig:graph_stats}
\end{figure}

\paragraph{Structural Shifts Triggered by Prompting Styles}
To better understand these performance variations, we examine the corresponding changes in reasoning graph topology across prompting conditions (Figure~\ref{fig:graph_stats}). It turns out that zero-shot prompting induces richer, more complex graph structures: graphs feature higher exploration density, greater branching and convergence, and a more diverse distribution of reasoning step counts (Figure~\ref{fig:step_cdf}). This suggests that, when not being influenced by extra demonstrations, the model engages in more adaptive and active exploration, revisiting, branching, and synthesizing reasoning steps.

In contrast, increasing the number of few-shot exemplars, especially in more verbose forms, systematically reduces both branching and convergence ratios, resulting in more linear graph architectures. The model appears to mimic the structure of the provided examples, limiting its capacity for active online reasoning. Notably, even a single demonstration can trigger a significant distributional shift toward shorter, more stereotyped reasoning chains.

These findings underscore the critical role of few-shot prompting styles in shaping RLM reasoning. Including no or a few extra demonstrations encourages flexible exploration and integration, which fosters more RLM's innate sophisticated reasoning graphs and leads to higher task performance. This motivates a shift in prompt engineering: effective demonstrations should balance informativeness with structural diversity, avoiding excessive \textit{unnecessary} context that suppresses the model's reasoning potential.

\subsection{Structural Signatures of Effective Reasoning}

\begin{table}[tb]
    \centering
    \begin{adjustbox}{width=1.00\linewidth}
    \renewcommand{\arraystretch}{1.30}
    {
    \fontsize{20.74}{24}\selectfont
        \begin{tabular}{c c >{\centering\arraybackslash}m{2.5cm} >{\centering\arraybackslash}m{2.5cm} >{\centering\arraybackslash}m{2.5cm} >{\centering\arraybackslash}m{2.5cm} c}
        \midrule
        Prompt Type & Acc (\%) & $\rho_E$ & $\gamma_B$ & $\gamma_C$ & $\ell$ & Mean Steps \\
        \midrule
        \multicolumn{7}{c}{\textsc{Llama-8B}$^*$} \\ 
        \midrule
        Zero-shot & 44.5 & 0.117 & 0.564 & 0.676 & 0.744 & 11.6 \\
        Concise & 41.2 & 0.065 & 0.252 & 0.426 & 0.947 & 8.6 \\
        Explanatory & 32.7 & 0.057 & 0.238 & 0.392 & 0.926 & 10.8 \\ \midrule
        \multicolumn{7}{c}{\textsc{Qwen-14B}$^*$} \\ \midrule
        Zero-shot & 51.8 & 0.122 & 0.612 & 0.719 & 0.716 & 15.8 \\
        Concise & 48.5 & 0.069 & 0.264 & 0.453 & 0.931 & 12.2 \\
        Explanatory & 45.3 & 0.061 & 0.243 & 0.420 & 0.946 & 14.2 \\ \midrule
        \multicolumn{7}{c}{\textsc{Qwen3-32B}} \\ \midrule
        Zero-shot & 56.2 & 0.188 & 0.835 & 0.760 & 0.444 & 19.0 \\
        Concise & 53.7 & 0.110 & 0.529 & 0.634 & 0.766 & 15.6 \\
        Explanatory & 50.1 & 0.069 & 0.283 & 0.449 & 0.931 & 18.0 \\ \midrule
        \end{tabular}
    }
    \vspace{-2mm}
    \end{adjustbox}
    \caption{Comparison of model performance and reasoning graph metrics across different model sizes and prompting paradigms. $^*$ denotes reasoning models that are officially distilled from \textsc{DeepSeek-R1}.}
    \label{tab:main}
\end{table}

To comprehensively reveal the relationship between reasoning structure and task performance, we present in Table~\ref{tab:main} a systematic comparison of graph-based metrics across multiple model scales and prompting paradigms. Several key patterns emerge that robustly distinguish effective reasoning.

\paragraph{Sophisticated Reasoning Graph Structure Drive Success.}
Across all models and prompt types, higher accuracy is consistently associated with richer reasoning graph structure: increased exploration density ($\rho_E$), higher branching ratio ($\gamma_B$), and greater convergence ratio ($\gamma_C$). Notably, larger models (e.g., Qwen3-32B) exhibit both the highest accuracy and the most complex graph structures, particularly under zero-shot prompting. This indicates that effective reasoning is achieved through a harder exploration (multiple attempts generation) and integrative convergence (synthesizing reasoning threads).

\paragraph{Prompt Constraints Induce Linearity and Impair Performance.}
Prompt types that impose stronger structural constraints consistently yield lower branching and convergence ratios, along with increased linearity ($\ell$). This shift toward more linear graph topologies is directly correlated with performance degradation. The effect is most pronounced in smaller models but persists even for larger architectures. These results again highlight the double-edged nature of few-shot demonstrations for RLM.

\paragraph{Quantitative Correlations.}
We extend the Pearson correlation analysis to all four graph-based metrics and observe that exploration density ($r=0.68$), branching ratio ($r=0.67$), and convergence ratio ($r=0.68$) each exhibit a strong positive association with accuracy with all significant at the $0.05$ level. These results indicate that denser, more exploratory and convergent reasoning paths are closely linked to model accuracy. Notably, these trends persist across all model scales and prompting regimes, highlighting the robustness and explanatory value of our structural framework.

In summary, these results provide compelling evidence demonstrating that reasoning graph analysis provides highly correlated and deep insights into the internal cognitive dynamics of reasoning language models.

\section{Conclusion}
This paper introduces a novel graph-based framework for analyzing reasoning output produced by reasoning large language models, offering quantifiable findings about how models \textit{organize} their thought flow under various factors. 
We first propose a reasoning graph toolkit that efficiently converts raw Chain-of-Thought tokens into analyzable graph structures. 
We then offer discovery regarding how various prompting styles may cast significant influence on RLM's internal reasoning structure and thus final performance.
We also provide strong evidence supporting that graph-level predictors strongly correlate with RLM problem-solving performance. 
These findings not only establish quantitative insights for future prompt engineering for reasoning models, but also provide a new structural perspective for evaluating reasoning quality beyond traditional metrics and interpreting how LLMs reason at a higher level.

\section{Limitations \& Future Work}

While our current graph-based analysis of RLMs focuses primarily on mathematical and coding tasks, extending this framework to broader domains, such as multi-modal or open-domain reasoning, may yield deeper insights into model behavior across varied scenarios and further inform our understanding of test-time scaling. In addition, the quantifiable structural metrics we propose provide a foundation for future research to explore more localized patterns and relational dynamics within reasoning graphs. We believe that ongoing work along these lines can contribute to a more comprehensive understanding and interpretability of large language models in practice.

\bibliography{anthology,custom}
\bibliographystyle{acl_natbib}

\newpage
\appendix

\section{Experimental Prompts}
\label{appendix:prompts}
\begin{figure}[tb]
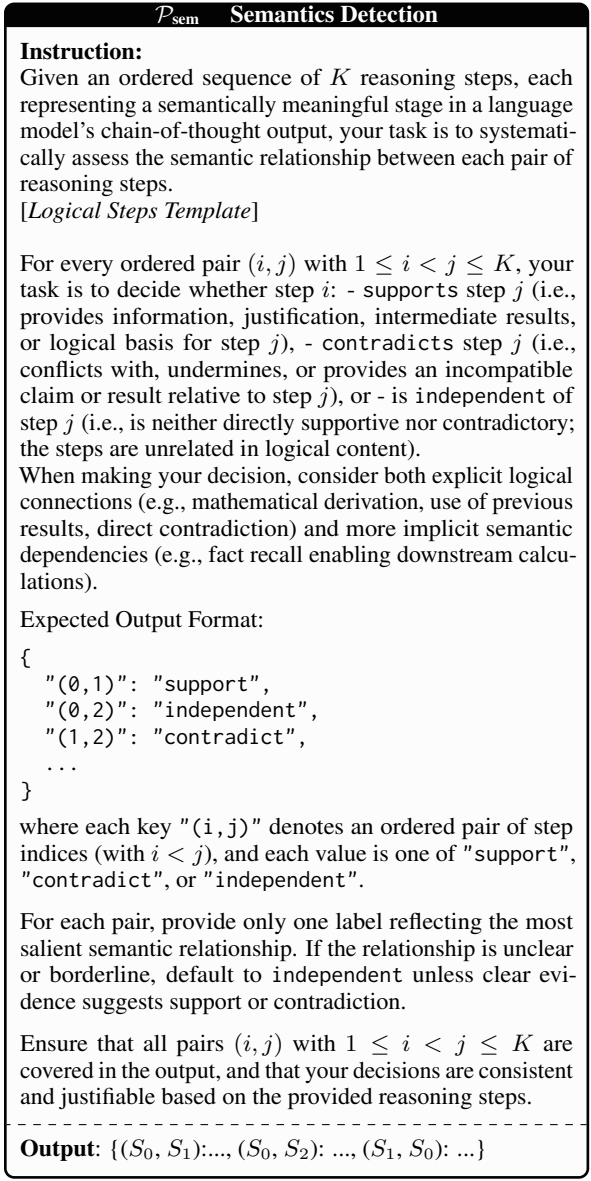

\centering

\begin{tcolorbox}[width=\linewidth, fonttitle = \small\bfseries, title=$\mathcal{P}_{\text{sem}}$ \quad Semantics Detection,colframe=gray!2!black,colback=gray!2!white,boxrule=1pt,boxsep=0pt,left=5pt,right=5pt,fontupper=\footnotesize, halign title = flush center]
\textbf{Instruction:}

Given an ordered sequence of $K$ reasoning steps, each representing a semantically meaningful stage in a language model's chain-of-thought output, your task is to systematically assess the semantic relationship between each pair of reasoning steps.

[\textit{Logical Steps Template}]\\

For every ordered pair $(i, j)$ with $1 \leq i < j \leq K$, your task is to decide whether step $i$:
- \texttt{supports} step $j$ (i.e., provides information, justification, intermediate results, or logical basis for step $j$),
- \texttt{contradicts} step $j$ (i.e., conflicts with, undermines, or provides an incompatible claim or result relative to step $j$), or
- is \texttt{independent} of step $j$ (i.e., is neither directly supportive nor contradictory; the steps are unrelated in logical content).

When making your decision, consider both explicit logical connections (e.g., mathematical derivation, use of previous results, direct contradiction) and more implicit semantic dependencies (e.g., fact recall enabling downstream calculations).

\vspace{0.5em}
Expected Output Format:
\begin{verbatim}
{
  "(0,1)": "support",
  "(0,2)": "independent",
  "(1,2)": "contradict",
  ...
}
\end{verbatim}
\noindent where each key \texttt{"(i,j)"} denotes an ordered pair of step indices (with $i<j$), and each value is one of \texttt{"support"}, \texttt{"contradict"}, or \texttt{"independent"}.

\medskip
\noindent For each pair, provide only one label reflecting the most salient semantic relationship. If the relationship is unclear or borderline, default to \texttt{independent} unless clear evidence suggests support or contradiction.

\medskip
\noindent Ensure that all pairs $(i, j)$ with $1 \leq i < j \leq K$ are covered in the output, and that your decisions are consistent and justifiable based on the provided reasoning steps.

\tcbline
\textbf{\color[RGB]{0,0,0}{Output}}: \{($S_0$, $S_1$):..., ($S_0$, $S_2$): ..., ($S_1$, $S_0$): ...\}
\end{tcolorbox}

\caption{Complete instruction ($\mathcal{P}_{\text{sem}}$) for detecting semantical relationship among all reasoning steps.}
\label{fig:sem-prompting}
\end{figure}

To facilitate robust and reproducible semantic analysis of RLM-generated reasoning traces, we design two explicit prompting templates, detailed documented in Template~\ref{fig:clu-prompting} and Template~\ref{fig:sem-prompting}. 

Template~\ref{fig:clu-prompting} guides the model to cluster reasoning units into coherent, higher-level reasoning steps. This step provides context-aware candidate instances of clustering of long chain-of-thought tokens.

Template~\ref{fig:sem-prompting} is used to extract the semantic relationship between every pair of reasoning steps. By explicitly labeling each pair as \texttt{support}, \texttt{contradict}, or \texttt{independent}, this template enables the later probabilistic estimation of interdependencies within the model's chain-of-thought.

\begin{figure}[tb]
\centering

\begin{tcolorbox}[width=\linewidth, fonttitle = \small\bfseries, title=$\mathcal{P}_{\text{clu}}$ \quad Logical Clustering ,colframe=gray!2!black,colback=gray!2!white,boxrule=1pt,boxsep=0pt,left=5pt,right=5pt,fontupper=\footnotesize, halign title = flush center]
\textbf{Instruction:} 

You are given a sequence of reasoning units, each representing a contiguous fragment from a language model's chain-of-thought (CoT) output. These units have typically been segmented using raw delimiters and may be overly fine-grained or fragmented for downstream analysis.

[\textit{Input Template}]
\vspace{0.5em}

Your task is to cluster consecutive reasoning units that are semantically connected, producing a concise and coherent set of higher-level reasoning steps. Each reasoning step should:
- Combine all units that express a single coherent sub-task, logical inference, or closely related set of thoughts. Aim to group together units that collectively advance the same intermediate goal or logical point.
- Ensure that each resulting reasoning step contains enough self-contained context to be analyzed independently, but avoid excessive merging that would result in overly broad or incoherent segments.
- Maintain the original sequential order of reasoning.
- Avoid splitting apart reasoning units that clearly belong to the same sub-problem or share strong contextual dependency.
- Use concise yet informative titles for each reasoning step, reflecting its main logical function or purpose (e.g., "Restate Problem", "Recall Known Facts", "Solve Equation", "Synthesize Solution", etc.).

\vspace{0.5em}
Expected Output Format:
\begin{verbatim}
{
  "s0": {"title": "...", "content": "..."},
  "s1": {"title": "...", "content": "..."},
  ...
}
\end{verbatim}
\noindent where each \texttt{"sX"} key indexes an ordered reasoning step, with an appropriate \texttt{"title"} summarizing its logical purpose and \texttt{"content"} containing the merged, cleaned reasoning text.

\vspace{0.5em}
\noindent Please ensure the output is structured, coherent, and well-suited for subsequent semantic analysis or graph-based modeling of the reasoning process.

\tcbline
\textbf{\color[RGB]{0,0,0}{Output}:} \{"$s_0$": \{"title":..., "content":...\},...\}
\end{tcolorbox}

\caption{Complete instruction ($\mathcal{P}_{\text{clu}}$) for clustering reasoning units into logical cohesive reasoning steps.}
\label{fig:clu-prompting}
\end{figure}

\section{Few-Shot Prompting Styles}
\label{appendix:few_shot_style}

Most existing research works reporting the performance degradation of RLM given few-shot prompting have not explicitly analyzed the structure of few-shot examples, while the concrete formulation of few-shot demonstrations could play a significant role on the behavior of language model. To isolate potential structural factors contributing to this performance degradation, in this paper, we introduce and analyze three distinct few-shot prompting style:

\noindent
i) \texttt{Minimal:} Minimal exemplars containing only problem statements and final answers, without intermediate reasoning steps or explanatory content.

\noindent
ii) \texttt{Concise:} Human-authored concise reasoning traces characterized by \textit{short}, linear progression from problem formulation to solution with minimal exploration.

\noindent
iii) \texttt{Explanatory:} RLM-generated \textit{long} reasoning sequences featuring extensive problem space exploration, iterative verification mechanisms, and explicit self-reflection.

\section{Implementation Details}
\label{appendix:impl}

\paragraph{Models}
We evaluate a range of reasoning LLMs, including DeepSeek-R1-distilled-Llama-8B, DeepSeek-R1-distilled-Qwen-14B, and \textsc{Qwen3-32B}~\cite{yang2025qwen3}. For conditional sampling ($P_{\text{LLM}}$), we adopt \textsc{DeepSeek-V3-0324}~\cite{deepseek2024v3}.

\paragraph{Hyper-parameters}
To ensure reproducibility, we set the generation temperature to 0 when producing reasoning chains (CoT) with RLMs. During logical clustering and semantics detection processes, we keep $\tau_r \sim [0.3, 0.7]$ to ensure sampling diversity.

\paragraph{Datasets}
All experiments are conducted on the GPQA-Diamond benchmark~\cite{rein2023gpqa}. To facilitate robust reasoning analysis and avoid potential training data contamination, we convert multiple-choice items into open-ended questions, requiring models to actively generate reasoning CoT as well as final answer rather than matching existing choices.

\section{algorithm}
\label{appendix:algorithm}
This appendix provides the detailed pseudocode for the core algorithmic components of our framework: ensemble-based clustering of reasoning units~\ref{alg:ensemble‐selection}, and adaptive sampling-based construction of the semantic dependency graph~\ref{alg:edge-construction}. These algorithms operationalize the methods described in the main text, clarifying the iterative processes and statistical aggregation techniques used to ensure robust, uncertainty-aware structure induction from RLM outputs.

\begin{algorithm}[tb]
\caption{Ensemble-Based Clustering of Reasoning Units}
\label{alg:ensemble‐selection}
\KwIn{Reasoning units $U$, clustering prompt $\mathcal{P}_{\text{cluster}}$, 
      number of samples $B$, temperature grid $\{\tau_1,\dots,\tau_B\}$}
\KwOut{Selected segmentation $C^{*}$}

\For{$b \leftarrow 1$ \KwTo $B$}{
    $C^{(b)} \sim P_{\text{LLM}}\!\left(S \mid \mathcal{P}_{\text{cluster}},\, U;\, \tau_b\right)$\;
    $F^{(b)} \leftarrow F\!\bigl(C^{(b)}\bigr)$\;
}
$C^{*} \leftarrow \arg\max_{b} F^{(b)}$\;
\Return $C^{*}$\;
\end{algorithm}

\begin{algorithm}[t]
\caption{Adaptive Sampling-based Semantic Edge Construction}
\label{alg:edge-construction}
\KwIn{Reasoning steps $S = (s_1,\ldots,s_K)$, prompt $\mathcal{P}_{\text{sem}}$, confidence threshold $\varepsilon$, maximum samples $R_{\max}$, thresholds $\tau_{\text{pos}}$, $\tau_{\text{neg}}$}
\KwOut{Adjacency matrix $A$, edge weights $W$}

Initialize counts: $C_{ij}(c) \gets 0$ for all $i < j$ and $c \in \{-1, 0, +1\}$ \;
$r \gets 0$\;
\Repeat{$\max_{i<j} \mathrm{SE}_{ij} \leq \varepsilon$ \textbf{or} $r \geq R_{\max}$}{
    $r \gets r+1$\;
    Sample $A^{(r)} \sim P_{\text{LLM}}(A \mid \mathcal{P}_{\text{sem}}, V; \tau_r)$ \;
    \ForEach{$i < j$}{
        $c \gets A^{(r)}_{ij}$\;
        $C_{ij}(c) \gets C_{ij}(c) + 1$\;
    }
    \ForEach{$i < j$}{
        \ForEach{$c \in \{-1, 0, +1\}$}{
            $\hat{p}_{ij}(c) \gets \frac{C_{ij}(c)}{r}$\;
        }
        $\mathrm{SE}_{ij} \gets \sqrt{ \frac{1}{r} \sum_{l \in \{-1, +1\}} \hat{p}_{ij}(l)(1 - \hat{p}_{ij}(l)) }$\;
    }
}

\ForEach{$i < j$}{
    $w_{ij} \gets \hat{p}_{ij}(+1) - \hat{p}_{ij}(-1)$\;
    \eIf{$w_{ij} \geq \tau_{\text{pos}}$}{
        $A_{ij} \gets +1$, $A_{ji} \gets -1$\;
    }{
        \eIf{$w_{ij} \leq -\tau_{\text{neg}}$}{
            $A_{ij} \gets -1$, $A_{ji} \gets +1$\;
        }{
            $A_{ij} \gets 0$, $A_{ji} \gets 0$\;
        }
    }
    $W_{ij} \gets w_{ij}$, $W_{ji} \gets -w_{ij}$\;
}
\Return $A$, $W$\;
\end{algorithm}

\section{Reference Step Length $\mu_{\text{ref}}$}
\label{appendix:mu_ref}

The ideal average step length $\mu_{\text{ref}}$ serves as a reference for the length-regularity term and is computed by dividing the total number of tokens $N$ by a target number of reasoning steps $K_{\text{target}}$. We set $K_{\text{target}} = \min(\max(3, \lceil \sqrt{M} \rceil), 30)$, where $M$ is the number of initial delimiter-based segments. This square-root heuristic with lower and upper bounds ensures $\mu\_{\text{ref}}$ adapts to different CoT lengths and discourages both over-segmentation and overly coarse steps, enabling a scale-invariant and task-agnostic regularization.

\section{Standard-Error Derivation for the Signed Edge Confidence}
\label{appendix:stderr}

Let a single LLM adjacency sample yield a label
\(c\in\{-1,0,+1\}\) for the ordered pair \((i,j)\), corresponding to
\texttt{contradict}, \texttt{independent}, and \texttt{support},
respectively.  Define the random variable
\[
Z \;=\;
\begin{cases}
+1,& c = +1,\\
-1,& c = -1,\\
\;\;0,& c = 0.
\end{cases}
\]
The \emph{signed edge confidence} is the empirical mean of \(Z\) over
\(R\) samples,
\[
w_{ij}
   \;=\;
   \frac{1}{R}\sum_{r=1}^{R} Z^{(r)}
   \;=\;
   \hat p_{ij}(+1)\;-\;\hat p_{ij}(-1),
\]
where
\(\hat p_{ij}(c)=\frac{1}{R}\sum_{r}\mathbf{1}[c^{(r)}=c]\).

Because \(\mathbb{E}[Z]=p(+1)-p(-1)\),
only the two informative labels contribute to the signed mean.
The variance of \(Z\) under the true distribution~\(p\) is
\begin{align*}
&\operatorname{Var}(Z)\\
\;&=\;
(+1)^2p(+1) + (-1)^2p(-1) + 0^2p(0)\\
\;&-\;
\bigl[p(+1)-p(-1)\bigr]^2 \\
\;&=\;
p(+1) + p(-1)
\;-\;
\bigl[p(+1)-p(-1)\bigr]^2.
\end{align*}
Replacing \(p(\cdot)\) with their empirical estimates and dividing by
\(R\) yields an unbiased standard-error estimator:
\[
\mathrm{SE}_{ij} = 
\sqrt{
    \frac{1}{R}\sum_{l\in\{-1, +1\}}\hat p_{ij}(l)\bigl(1-\hat p_{ij}(l)\bigr)
}
\]
The independent label \((l=0)\) contributes neither positive nor
negative mass to \(Z\); its influence is expressed implicitly via the
complement \(1-p(+1)-p(-1)\).  A full multinomial variance expression
would add a term \(-2\,p(+1)p(-1)\), whose magnitude is
\(O(p(+1)p(-1))\) and empirically negligible for our edge-sparse
setting. Omitting this term simplifies the estimator while preserving
the accuracy required for the adaptive stopping criterion

\end{document}